\documentclass[a4paper]{article}

\usepackage[english]{babel}
\usepackage[utf8x]{inputenc}
\usepackage[T1]{fontenc}
\usepackage[affil-it]{authblk}
\usepackage{etoolbox}
\usepackage{lmodern}

\makeatletter
\patchcmd{\@maketitle}{\LARGE \@title}{\fontsize{14}{19.2}\selectfont\@title}{}{}
\makeatother

\usepackage[a4paper,top=3cm,bottom=2cm,left=3cm,right=3cm,marginparwidth=1.75cm]{geometry}
\usepackage[english]{babel}
\usepackage[autostyle]{csquotes}

\usepackage{amsmath}
\usepackage{graphicx}
\usepackage[colorinlistoftodos]{todonotes}
\usepackage[colorlinks=true, allcolors=blue]{hyperref}

\date{}
\newcommand{\ignore}[1]{{}}
\newcommand{\grouptext}[1]{\begingroup\leftskip2em\rightskip\leftskip{#1}\par\endgroup}
\newcommand{\hashtag}[1]{\textbf{\#{#1}}}
\newcommand{\hashtagT}[1]{\texttt{\#{#1}}}
\newcommand{\pp}{$^+$}
\newcommand{\nn}{$^-$}

\title{\textbf{Using Linguistic Cues for Analyzing Social Movements}}
\author{Rezvaneh Rezapour}
\affil{University of Illinois at Urbana-Champaign \protect\\
School of Information Sciences \protect\\ \{rezapou2@illinois.edu\}}
\begin{document}
\maketitle

\begin{abstract}
With the growth of social media usage, social activists try to leverage this platform to raise the awareness related to a social issue and engage the public worldwide. The broad use of social media platforms in recent years, made it easier for the people to stay up-to-date on the news related to regional and worldwide events. While social media, namely Twitter, assists social movements to connect with more people and mobilize the movement, traditional media such as news articles help in spreading the news related to the events in a broader aspect. In this study, we analyze linguistic features and cues, such as individualism vs. pluralism, sentiment and emotion to examine the relationship between the medium and discourse over time. We conduct this work in a specific application context, the ``Black Lives Matter'' (BLM) movement, and compare discussions related to this event in social media vs. news articles.   
\end{abstract}

\section{Introduction}

Micro-blogging platforms such as Twitter are central venues of conversation and arguments. Social movements (SM) aim to raise awareness on topics and events related to social issues and engage people in creating, changing, or removing policies, among other goals. The purpose, structure, and functioning of social movements may differ across time, issues, and cultures. Before the emergence of the internet, SM relied  on social gatherings, street actions, the printing press, and other broadcast media, including graffiti, radio, and TV. Online-based media, such as emails, forums, and social networking sites, complemented and enhanced the scope, reach, and impact of traditional media, and sometimes even fulfilled other functions, e.g., by using hashtags and mentioning powerful or influential online users \cite{bonilla2015ferguson,sandoval2014towards,tufekci2014medium}. While traditional media was limited in sense of allowing  people to actively engage or participate in movements, World Wide Web, emails, forums, and social media platforms helped activists to connect with more people and enhance their activities
\cite{sandoval2014towards}.

In the recent years, due to the rise of attention 
to social media, various activists, and organizations 
have brought the movements to this platform to 
(1) establish a better and more organized movement, 
(2) reach more audience online, and 
(3) inquire the attention of a younger generation. 
Easier access to the internet and electronic devices 
and the familiarity of different levels of society 
with the web and namely social media 
created a different situation in recent years 
to be active and engaged in political and social matters. 
One notable example of such situation is 
the Tahrir Square demonstration in Egypt. Throughout the years Egyptian government invested an enormous amount of resources 
to ``put PC in every home''
\cite{kamel2014egypt, kidd2016social}. 
The wise use of technologies and the availability 
of the resources helped people of this country 
in communicating better and more efficient 
which assisted in mobilizing the movement.

Moreover, the popularity of social media and micro-blogging has introduced hashtag activism in the recent years. 
Hashtags are powerful memes that are used for various reasons, 
from marketing to social movements and advocacy works 
\cite{guo2014tweeting}. 
Hashtag activism was first introduced for referring to 
``wall street protest'' by Guardian. 
Nowadays, most of the online SM are accompanied by specific hashtags which are created to support or globalize the movements. 
Using hashtags in social media such as Twitter makes the feed accessible to a wide range of people with the same interest and focus. On e example of such use is the Black Lives Matter movement (BLM), also known as \hashtagT{blacklivesmatter}, 
which received comparably wider attention (compared to previous movements) due to leveraging a 
systematic use of social media and hashtags
\cite{bonilla2015ferguson}. 

While social media, namely Twitter, assisted social movements to connect with more people and engage the society in the movement, traditional media such as news articles helped in spreading the news related to the events \cite{gamson1992talking}. Based on previous studies, news articles were found neutral in their tones while discussing controversial issues. However, regarding a major event such as social movements there have been limited studies to analyze and compare traditional media vs. social media data. 
This paper aims to fill this gap by analyzing linguistic features and cues, such as individualism vs. pluralism, emotion, language dimensions and sentiment in social media vs. news articles to examine the relationship between the medium and discourse over time. We conduct this work in a specific application context, the ``Black Lives Matter'' (BLM) movement, and compare discussions related to this event. 


Specifically, in this paper, we aim to analyze and discuss the following questions:
\begin{itemize}
\item How sentiment is correlated with time focus in both tweets and news articles?
\item How anxiety, sadness and anger vary in the two datasets?
\item Is the distribution of positive and negative emotions different across two datasets?
\item Can we observe any changes in the usage of words representing individualism and pluralism in tweets vs. news articles? 
\item Are there any differences between language dimensions in the two datasets?
\end{itemize}

As our analysis shows, people use more ``we'' and ``our'' in social media  when a major incident takes place in society representing an increase in pluralism. In addition, we found that both emotion and sentiment in language were significantly influenced by the major events in both news and Twitter. Moreover, we found a significant correlation between future and positive sentiment in tweets representing hope in user-generated texts. Finally, our analysis highlights that people connect and at least virtually take part in the movement.

\section{Background and Related Works}
\subsection{Online Movements and Social Change}

The notion of SM in social media has been widely studied 
in various domains such as political science, 
social psychology and sociology. 
The studies in this realm can be divided into three groups: 
``optimists'', ``pessimists'' and ``ambivalent'' 
\cite{kidd2016social}. 
As the names of the three categories show there are 
distinct beliefs about social media between scientists. 
While some believe that social media is tremendously important 
in mobilizing a movement (online movements), 
others find it naive to rely on online activism 
to win a battle against higher powers. 
The latter group believes that SM is only successful 
when people take action in their hands (offline movement) 
\cite{kidd2016social}. 

The positive opinion about the influence of social media 
comes from the movements that productively used this tool. 
One notable example of these movements is the 
{\em black lives matter} and {\em Ferguson} movements 
to raise the awareness of the American people about 
discrimination and the importance of equality and race diversity 
in the US \cite{bonilla2015ferguson}. 
These movements used social media to increase their influence as well as their public engagement. Moreover, they leveraged social media for decision making and getting more individuals to participate in various social events. 
Sandoval-Almazan and Gil-Garcia \cite{sandoval2014towards} 
studied three different cases of SM in Mexico and found that 
different SM uses social media, mobile phones and online technologies to counter fight the governmental action in suppressing the movements and in better mobilizing the movements 
to keep their participants active and motivated. 
On the other hand, numbers of researchers argue that the usage of social media or being active online does not necessarily result in a huge victory. Lack of guidance and an enormous amount of information online can easily perish the participants and disrupt the movements \cite{bonilla2015ferguson,kidd2016social}. 
In addition, social media were found to be used 
as a tool in the hand of authoritarian or democratic governments 
to track the move of activists and counter act their actions if/as needed \cite{bonilla2015ferguson,kidd2016social}. 
As an example, during the protests in Iran in 2009, 
the government filtered social media in designated regions 
for demonstration and stopped the mobile phone signals 
to diverge the movement \cite{christensen2011twitter}. 
While such examples highlight the negative use of social media and its influence on SM, the positive impact of such platforms can not be neglected as discussed in the next sections. 

\subsection{Hashtags and Social Movements}

Hashtags are used as a means of opening to social activism. 
Different studies found using hashtags as significant parts 
of social movements 
\cite{guo2014tweeting,kidd2016social,tsur2012s,zhang2015modeling}. 
One of the most influential online movements in the recent years 
took place after a young man was killed by a police officer in Ferguson. As shown in their study, Bonilla and Rosa \cite{bonilla2015ferguson} found social media and the usage of hashtags as significant parts of this social movement and showed that the usage of \hashtagT{Ferguson} brought attention to this incident inside and outside the US. 

Leveraging a unique and universal language to connect to people is one important factor to have a successful campaign (movement)
\cite{bonilla2015ferguson}. 
Hashtags are capable of providing such communication 
between all the individuals sharing the same interest. 
For instance, using smart hashtags such as \hashtagT{MarchYoSoy132} 
on Twitter and Facebook, in addition to other online technologies, 
in Mexico assisted people not only in finding and 
tracking the movement and demonstrating against the governmental power, it also provided an opportunity to disseminate correct information 
\cite{sandoval2014towards}. 
In the case of Mexico and Ferguson or similar movements, 
using a wise hashtag gave this opportunity to the campaigns 
to become a worldwide trend and bring attention 
to a social justice issue in a remote area. 
Obviously, ``trending'' a  movement/hashtag and its worldwide recognition will bring attention to the event which may force governments to act faster, address the issue, and enhance the situation in a shorter period of time. 
Social movements, in some cases, are affiliated with other campaigns and movements since (1) they are a part of bigger active campaigns, or (2) some similar incidents took place previously 
which are typically connected to the new ones, i.e. \hashtag{takeaknee}, \hashtag{Ferguson}, and \hashtag{blacklivesmatter}. Using hashtags make it possible for social movements and activists to link different incidents with the same foci throughout the time \cite{bonilla2015ferguson,tsur2012s}.

Moreover, while using hashtags can simplify the information retrieval Bonilla and Rosa \cite{bonilla2015ferguson} found that 
some popular hashtags are used to bring the attention of 
social media users to unrelated feeds such as advertisements and misleading information. Analyzing the source and content, namely verifying the users and studying the informations in tweets such as the contexts and URLs are among useful pointers for reducing  false positive inputs and assisting the users to navigate the trustworthy sources \cite{bonilla2015ferguson,guo2014tweeting}.  

Social media as a medium and tool is capable of connecting and mobilizing events and people and keeping social movements loud and active. Such archival power is missing in traditional media such as news channels, and radio. For a long time, especially in the absence of social media such as Twitter, traditional media has been used as a platform for studying, analyzing and discussing political and social issues \cite{gamson1992talking,benford2000framing,rohlinger2013media}. Regarding social movements, while traditional media has been used to ``mobilize a broader population to action'', internet ``provides a means through which movement groups can communicate with an audience directly'' \cite{rohlinger2013media}.  This paper aims to study the linguistic cues in both these platforms, namely media and social media, to examine the relationship between the medium and discourse over time on the subject of social movements. while the result of this paper can help in comparing the two mediums, it will also bring the attention on the impact of powerful sources such as societal events on  people's behavior, emotion and cognition \cite{rezapour2017classification}. 

 \section{Data}
Black Lives Matter and its related hashtag started in 2013. It was widely recognized in 2014 after the death of two African American, Freddie Gray, and Michael Brown, by the police. 
The timeline of this study is focused on the period between June 2014 and June 2015 
to capture changes in discourse in both Twitter and news articles. 
It is important to note that during this time 
numbers of important events took place in the US which significantly influenced the BLM movement.
The events are listed in Table~\ref{tab:events}.

\begin{table}[t]
\caption{List of significant events between June 1st, 2014 and June 1st, 2015}
\label{tab:events}
\centering
{\footnotesize
\begin{tabular}{rl}
\hline
Date & Events \\
\hline
JUL 17, 2014			& Death of Eric Garner \\ 
AUG 9, 2014				& Shooting of Michael Brown \\ 
AUG 9 - 25, 2014		& 1st wave of Ferguson protest \\
NOV 25, 2014			& Darren Wilson non-indictment \\
NOV 24 - DEC 2, 2014	& 2nd wave of Ferguson protest \\
DEC 3, 2014				& Daniel Pantaleo non-indictment \\
APR 4, 2015	 			& Shooting of Walter Scott \\
APR 12, 2015			& Death of Freddie Gray \\
\hline
\end{tabular}}
\end{table}

Regarding the tweet dataset, we used the BLM Dataset, provided by Center for Media and Social Impact \cite{freelon2016beyond}. This dataset consists of 40 million tweets related to the movements collected from June 1st, 2014, to June 1st, 2015. To reduce the complexity of computation, we sampled 500 tweets per day and ran the tweet IDs through the Twitter API. The final sample set of tweets consists of around 175K tweets in total. 

To collect the news articles, we used Lexis-Nexis Academic\footnote{http://www.lexisnexis.com/hottopics/lnacademic/}.
We first created and tested various queries to capture the articles related to the BLM with the least amount of noise and unrelated articles. 
After reviewing a random sample of news articles,
we found the following query the most effective one: 

\grouptext{
\noindent
\texttt{``(``black lives matter'') OR (``blacklivesmatter'') OR (``\hashtag{blacklivesmatter}'') OR (``ferguson'' AND (``blackmen'' OR ``blackman'' OR ``black person'' OR ``black people'' OR ``gunman'') AND ``shooting'')''}
}

The reason for choosing this query was that using ``Ferguson'' alone, resulted in getting articles related to athletic activities and basketball or baseball articles of the city (Ferguson). 
To minimize this noise, we added another word to the query as ``black men'' and ``shooting'' 
to retrieve articles suitable for this study. 
Using the final query, shown above, we collected around 2,866 news articles after removing the duplicates.

\section{Methodology}
\subsection{Preprocessing}
Before starting the analysis, both datasets were first processed and cleaned. 
Tweets can be highly noisy and follow unconventional
spelling schemes. Therefore, we preprocessed the tweet data with
replacing all URLs with the tag ``URL''; replacing all
usernames with ``USER'', removing or limiting repetitions of the same latter to two consecutive occurrences (e.g., changing ``gooooood'' to ``good''), 
and removing hashtag symbols. After that for both news dataset and tweets, 
we removed the punctuations and numbers and lowercased the words. We also removed the stop words to reduce the dimension of words especially in news articles.
\subsection{Part of Speech Tagging}
To analyze the news datasets, we used python NLTK library to tokenize the news articles and then tagged the tokens with their respective Part of Speech (POS). 
For tweets, we used a tokenizer and POS tagger suitable
for Twitter \cite{owoputi2013improved}. 
\subsection{Individualism vs. Pluralism}
When an event takes place in society, people who support that event tend to be a part of the movement and show their participation by posting on social media. Generally, individuals mostly use ``I'' and ``my'' in their everyday life to describe the events or express their opinions \cite{campbell2003secret}. However, to show their participation, people use more ``we'' and ``our'' to show their involvement in the process of change
or movement. To study this phenomenon, after tagging the part of speech, we extracted the number of ``I, we, you, your, yours, our, ours, and etc'' in the text. 

\subsection{Sentiment Analysis}
To analyze the sentiments of both tweets and news articles, 
we used MPQA subjectivity lexicon developed by Wiebe and
colleagues \cite{wilson2005recognizing}. The lexicon consists of 8,222 negative, positive and neutral words and their POS. To tag the tweets or news articles with their polarity, after tokenizing the data and tagging each token with its POS, we extracted the terms (with respect to their POS) which coincided with the MPQA lexicon entry. We calculated and counted the aggregated numbers of
positive, negative and neutral tokens per tweet or news article and tagged them with the largest polarity class. For tweets, In addition to the words in MPQA lexicon, we extracted and annotated top hashatgs from the tweets and added them to increase the efficiency of sentiment analysis \cite{rezapour2017identifying}

\subsection{Emotion Analysis, Time, and Language Dimensions}
In addition to sentiment, we decided to study the change in emotion and language of both tweets and news articles in the mentioned time span. For both of these feature sets, we used Linguistic Inquiry and Word Count tool known as LIWC \cite{pennebaker2015development}. This tool is broadly used for capturing various words categories based on its embedded lexicons. 

To study emotions, we extracted positive and negative emotions, 
as well as sadness, anger, and anxiety from tweets and news articles using LIWC. Using this tool we also extracted the time focus of each text or tweet. The times are extracted with respect to the verb or adverbs used in the sentences. In addition, to study the language of each text, we used Clout, Analytic Thinking, Authentic, and Emotional Tone. 
\begin{itemize}
\item \textit{Analytic thinking} (analytic): 
``high values reflect a formal, logical and hierarchical thinking, 
while the low values reflect more informal, personal and narrative thinking.''
\item \textit{Clout}: 
``high values suggest that authors are speaking with high expertise and confident, 
while the low numbers suggest a more tentative, humble and anxious style.''
\item \textit{Authenticity} (authentic): 
``high numbers are associated with a more honest, personal and disclosing text, 
while low numbers suggest more guarded, distanced form of discourse.''
\item \textit{Emotional tone}: 
``high values reflect more positive and upbeat style,
while lower values reflect high anxiety, sadness or hostility.''
\end{itemize}

\subsection{Semantic Networks}
Semantic networks is used to gain a more comprehensive insight on the discourse and the development of dialogue around an issue \cite{diesner2015social}. 
News articles retrieved from Lexis Nexis are accompanied with various types of index terms called meta-data which feature the topics of the news articles. The terms come with a threshold number that shows how much of the content of each article is related to the tagged index. The meta-data can be related to entity types such as subject, language, region, person and so on. 
In this study, we focused on the subject terms to study the change in discourse in news articles. 

To construct the semantic networks we first extracted the index terms with at least 75 percent threshold per article and then developed the semantic network Figure~\ref{fig:networks}. The nodes in these networks represents the most salient information and the edges represent the co-occurrences of the terms within seven words window of the news dataset. 
We used ConText \cite{diesner2014context} to extract the meta-data  and construct the network. 

\section{Analysis and Results}

\subsection{Sentiments Analysis}

\begin{figure}[t]
\centering
\includegraphics[width=8.5cm]{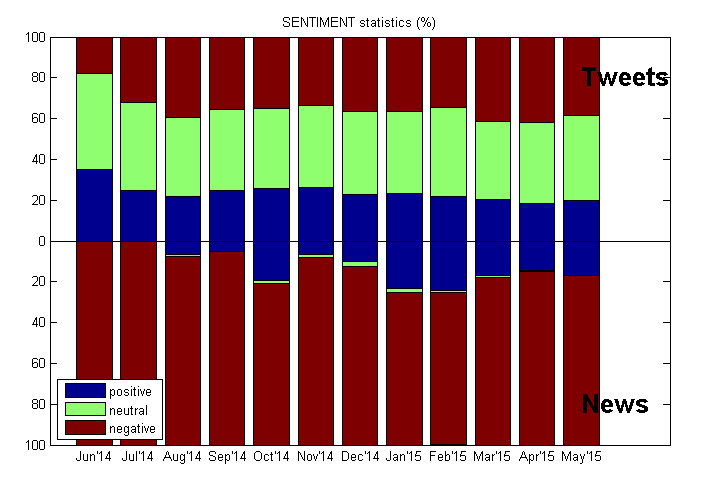}
\caption{Sentiments analysis statistics in tweets and news article}
\label{fig:sentiment}
\end{figure}

Figure~\ref{fig:sentiment} shows the percentage of the sentiment values (positive, neutral and negative) presented in tweets and news articles. As the result shows, news articles carry more negative sentiment values compared to the tweets. The sentiment polarities in tweets are more evenly distributed though the negative and neutral sentiments are more prevalent than the positive sentiment. Since the tweets are user- generated, they highlight the personal opinions. Therefore, the information found on tweets are on various aspects of the movement from furious ranting of a victim of a shooting to praying and hoping for the success of the movement. As a result, the sentiment values found in tweets are more dispersed. News articles usually present details of a particular event and include group discussions, analysis, and interviews by journalists and experts. 
The sentiment values are, therefore, negative since the discussion on the related events are usually as such. 

For observation on the results of tweets, we found that the sentiment values changed after particular events. For instance, the negative sentiment kept increasing and the positive sentiment kept decreasing since July 2014 after the death of Eric Garner. 
The two sentiments followed the same trends through August due to
the death of Michael Brown and the initial Ferguson protests during the same month. The two sentiment values remained roughly the same since then until March of 2015 even though there were two main events related to the non-indictment of Darren Wilson and 
Daniel Pantaleo during November and December of 2014.
It is possible that non-death-related event had less impact on the sentiment since we did not observe any raise in negative sentiments. However, these events were sufficient to maintain the overall sentiments. With the protests at the end of March and the Walter Scott case in early April 2015, a similar change of sentiment values can be observed in tweets dataset.

\subsection{Sentiments and Time-foci Analysis}

\begin{table*}[t]
\caption{$p$-value between sentiments and time-foci. 
Significant results at 0.05 are shown in bold numbers.}
\label{tab:timeVSsentiment}
\centering
\begin{tabular}{r|ccc|ccc}
focus {\textbackslash} sentiment & positive & neutral & negative & positive & neutral & negative \\
\hline
past-focus		& .310	& .999	& .482	& .230	& .511	& .227 \\
present-focus	& .177	& .565	& .491	& .056	& .164	& \textbf{.049} \\
future-focus	& \textbf{.021}	& .661	& .088	& .050	& .471	& .052 \\
\hline
source			& & tweets & & & news & \\
\end{tabular}
\end{table*}

We further analyzed sentiment values with respect to various time-foci in the studied texts. We categorized the time-focus of each text data to either past-focus, present-focus, and future-focus and computed the correlation between sentiments and the time-foci using Pearson's correlation at $p$-value 0.05.
The results are shown in Table~\ref{tab:timeVSsentiment}.
In tweets, future-focus and positive sentiment are significantly correlated. In fact, when tweets consist of future-focus, they are mostly about hope and pray for a better future.
Examples of tweets that show positive sentiment and future-focus are as follows.

\grouptext{
\noindent
\texttt{``Gov. Jay Nixon speaking now. `Our hope and expectation in the coming weeks is that peace will prevail...' \hashtag{ferguson}''}, 
\textit{anonymous1}
}

\grouptext{
\noindent
\texttt{``Tonight we \hashtag{standtogether} we will stand to \hashtag{remember} to \hashtag{change}. 
Will u stand with us?''}, 
\textit{anonymous2}
}

In news articles, as expected, present-focus and negative sentiment are significantly correlated. 
This is because the texts are focused on the current event, including interviews and analysis, reflecting the event at the moment which, again, are usually unpleasant.
Examples of texts in news articles that show negative sentiment and present-focus are as follows.

\grouptext{
\noindent
\texttt{``... we worry about losing our lives with any interaction we have with police.''}, 
stripped text given by a freelance writer from an article on August 22, 2014
}

\grouptext{
\noindent
\texttt{``It's easier to focus on distracting details of a shooting, or political battles between a governor and a prosecutor, and safe in our suburban enclaves, turn the page on Ferguson and pretend it was all a hazy hallucination.''}, 
article published by St. Louis Post-Dispatch Inc. on August 24, 2014
}

\subsection{Individualism and Pluralism Analysis}

\begin{figure}[t]
\centering
\includegraphics[width=10cm]{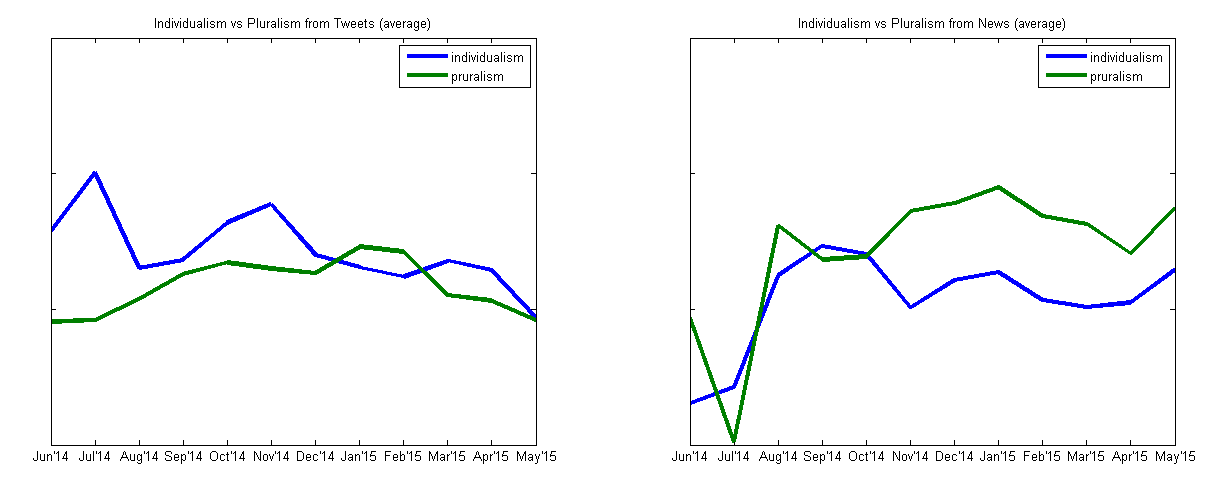}
\caption{Individualism and Pluralism in tweets and news article}
\label{fig:grpfocus}
\end{figure}

As mentioned earlier, we believe that people connect to movements and events by expressing their feelings and changing their language by using pronouns such as ``we'' instead of ``I''.
Figure~\ref{fig:grpfocus} shows statistics of words that
represent individualism, e.g., \textit{I} or \textit{me}, 
and words that represent pluralism, e.g., \textit{we} or \textit{us}, in both tweets and news articles.
We do not observe any clear pattern of change in their usage
in news articles. 
Overall, news article tends to be more formally written. They mostly consist of plural words rather than \textit{I}
in most articles. Observing news articles showed that \textit{I} was mostly used in interviews of individuals across news articles.

On the other hand, we observe that pluralism increased after important events in tweets. For instance, this trend was observed during Eric Garner incident in July 2014. 
The trend was observed again during Ferguson protests from August 
through November of 2014 where pluralism increased constantly
and then slightly dropped afterward. Similar change is shown again during Walter Scott and Freddie Gray
cases in April 2015. These observations imply that people tend to, collectively, show they are a part of the movement or are influenced by such incidents, especially during the protests.   
The usage of pluralism are prevalent in posts that involves widespread 
impact or posts that are more directed to society in general such as the following:

\grouptext{
\noindent
\texttt{``You think we WANT to protest? Nah. We wanna live. We protest because we are being slaughtered. \hashtag{ferguson}''}, 
\textit{anonymous3}
}

However, the use of individualism are more prevalent on posts that
present opinions or emotions at the moment.
Examples of such posts are as follows.

\grouptext{
\noindent
\texttt{``I'm in tears sitting here at my desk \hashtag{ferguson}''}, 
\textit{anonymous4}
}

\grouptext{
\noindent
\texttt{``I can't believe what I'm reading... R.I.P \hashtag{AntonioMartin} \hashtag{policebrutality} \hashtag{civilrights} \hashtag{BLACKLivesMatter}''}, 
\textit{anonymous5}
}

\subsection{Emotions Analysis}

\begin{figure}[t]
\centering
\includegraphics[width= 10 cm]{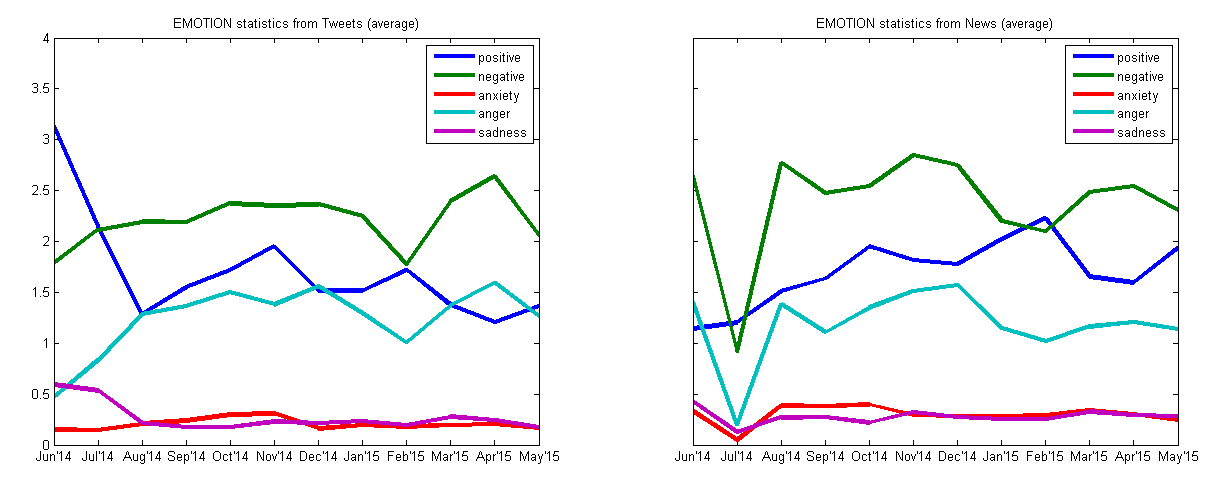}
\caption{Emotions statistics in tweets and news article}
\label{fig:emo}
\end{figure}

\begin{table*}[t]
\caption{$p$-value between analyzed emotions. 
pos, neg, anx, ang and sad represent 
positive, negative, anxiety, anger and sadness emotions, respectively.
$+$ and $-$ signs denote positive and negative correlation, respectively.
Significant results at 0.05 are shown in bold numbers.}
\label{tab:emoVSemo}
\centering
{\footnotesize
\noindent\makebox[\textwidth]{
\begin{tabular}{r|cccc|cccc|ccccc}
emotions & neg & anx & ang & sad & neg & anx & ang & sad & pos & neg & anx & ang & sad \\
\hline
pos	& \textbf{.041}\nn	& .557\nn	& \textbf{.001}\nn	& \textbf{.001}\pp	& .515\pp	& .445\pp	& .399\pp	& .642\nn	& .080\nn	& .621\nn	& .534\nn	& .789\nn	& .275\pp \\
neg	& -			& .152\pp	& \textbf{.001}\pp	& .210\nn	& -			& \textbf{.001}\pp	& \textbf{.001}\pp	& .009\pp	& .876\pp	& .389\pp	& .601\pp	& .488\pp	& .647\nn \\
anx	& -			& -			& .093\pp	& .097\nn	& -			& -			& \textbf{.003}\pp	& .068\nn	& .282\pp	& .145\pp	& .094\pp	& .201\pp	& .983\pp \\
ang	& -			& -			& -			& \textbf{.001}\nn 	& -			& -			& -			& \textbf{.014}\pp 	& .098\pp	& .175\pp	& .255\pp	& .220\pp	& .541\nn \\
sad	& -			& -			& -			& -			& -			& -			& -			& -			& \textbf{.002}\nn	& .145\nn	& .109\nn	& .181\nn	& .663\pp \\
\hline
source	& \multicolumn{4}{c|}{tweets} & \multicolumn{4}{c|}{news} & \multicolumn{5}{c}{tweets (column) vs news (row)} \\
\end{tabular}}
}
\end{table*}

Figure~\ref{fig:emo} shows statistical analysis of emotions shown in tweets and news articles.
Overall, we observed that in most categories different emotional values shifted before and after death-related events. For instance, after the death of Eric Garner and Michael Brown between July and August, the positive emotion in tweets decreased and the negative emotion increased. Regarding the news articles, we observed a sharp increase in the negative emotion during the same period. By the end of April where the incidents of Freddie Gray and Walter Scott took place, we observed similar trends in both tweets and news articles. Regarding anger, we observed similar trend to the negative emotion in both
tweets and news articles.

Next, we performed correlation analysis between the emotion categories within and between both news articles and tweets using Pearson's correlation at $p$-value of 0.05. 
The results are shown in Table~\ref{tab:emoVSemo}.

Overall, we found that sadness in tweets and positive emotion in news articles are significantly negatively correlated. 
Correlation between other emotional categories in tweets and in news articles were not significant. For news articles, we found no surprise in the correlation results. As expected, we found that 
(1) negative emotion and anxiety, (2) negative emotion and anger,
and (3) negative emotion and sadness are significantly correlated.

On the other hand, in tweets, we observed the expected significant negative correlations between positive emotion and negative emotion and between positive emotion and anger.
We also observed the expected significant correlations 
between negative emotion and anger. Surprisingly, we observed that there is a significant correlation between positive emotion and sadness instead of the expected negative correlation between them.
By analyzing texts of tweets with both positive emotion and sadness, indeed, we found two possible explanations. 
First, these texts are mostly about hope and wish.
However, since the texts are related to the BLM movement and
its events, they are likely to include words that express sadness.
Examples of such tweets are as follows.

\grouptext{
\noindent
\texttt{``There are tears, and there is joy, and hope here. This is community. This is love. \hashtag{Ferguson} \hashtag{BlackLivesMatter}''}, 
\textit{anonymous6}
}

\grouptext{
\noindent
\texttt{``Sorry for seeming flippant reference to \hashtag{Ferguson}. I think it's heartbreaking and brave what is going on there. Love and hope to you.''}, 
\textit{anonymous7}
}

The other explanation is simply sarcasm which is shown in the following examples.

\grouptext{
\noindent
\texttt{``and I will voice my sincere disappoint. \hashtag{cantwealljustgetalong} \hashtag{BlackLivesMatter} lol''}, 
\textit{anonymous8}
}

We also found that sadness and anger are significantly negatively correlated in tweets data. For this findings, we could not find an appropriate tweets example that satisfies this result
and, thus, we could not provide any explanation for this case.
However, it is likely that the lack of anger expression in tweets with sadness or vice versa resulted in such negative correlation between the two emotions.
Nevertheless, there are tweets that express both sadness and anger; however, such tweets are not prevalent in our dataset.
Tweets with both sadness and anger are mostly related to death-involving events such as the shooting of Michael Brown and less about the protest-related events. 
In case of anger, our observation shows that most of the tweets that express anger usually involve using swear words, and the tweets are mostly directed towards police in general. 

\begin{figure}[t]
\centering
\includegraphics[width=10 cm]{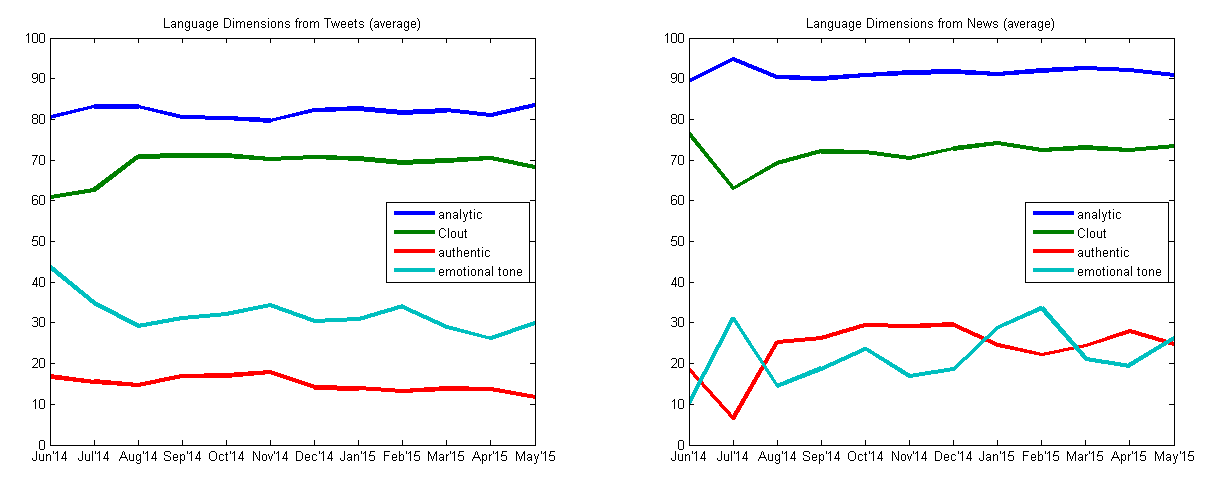}
\caption{Language Dimensions statistics in tweets and news article}
\label{fig:ldim}
\end{figure}

\begin{figure*}[ht!]
\centering
{\scriptsize
\begin{tabular}{@{}c@{}c@{}}
\includegraphics[height=6.3cm]{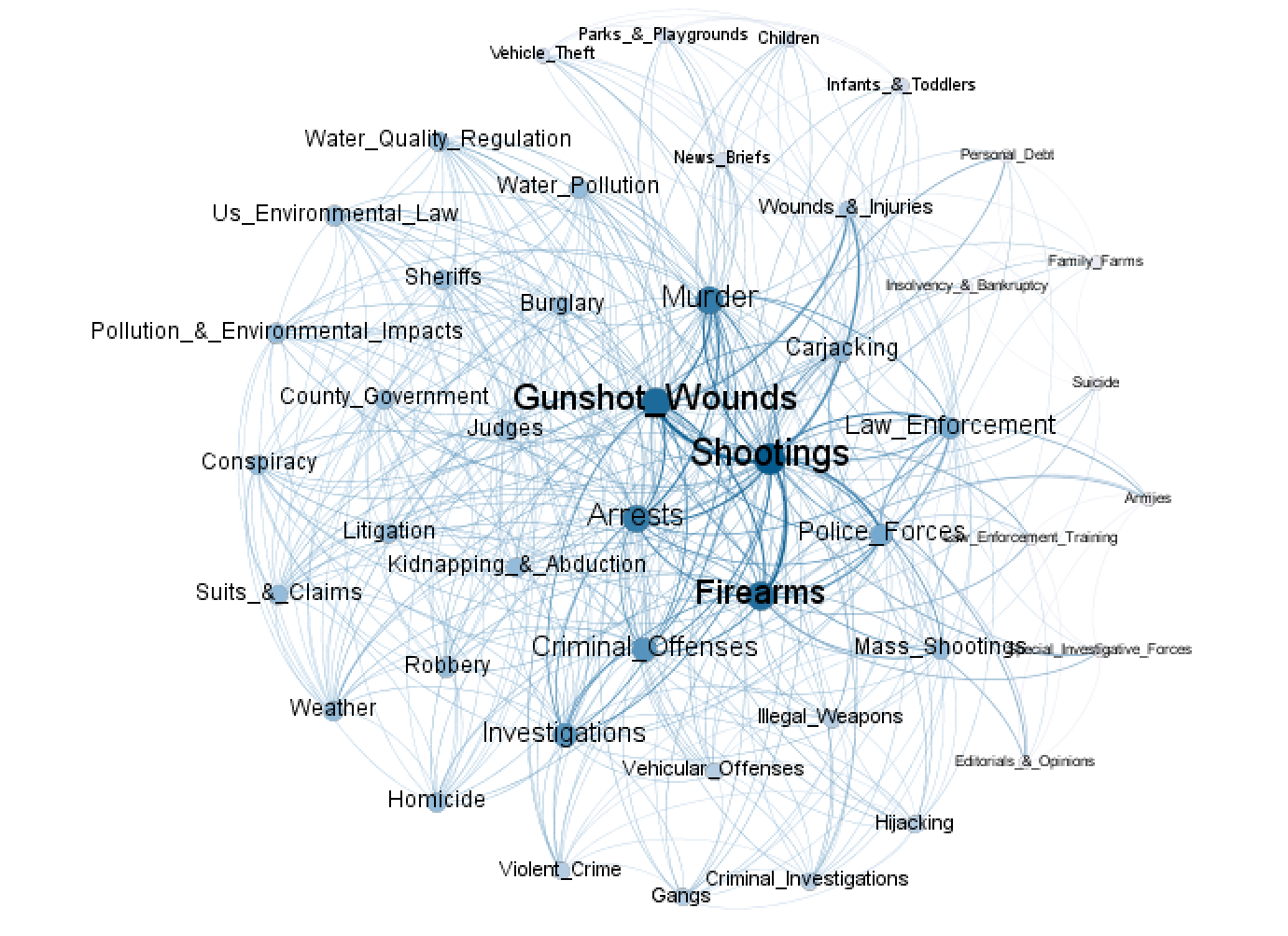} & \includegraphics[height=6.3cm]{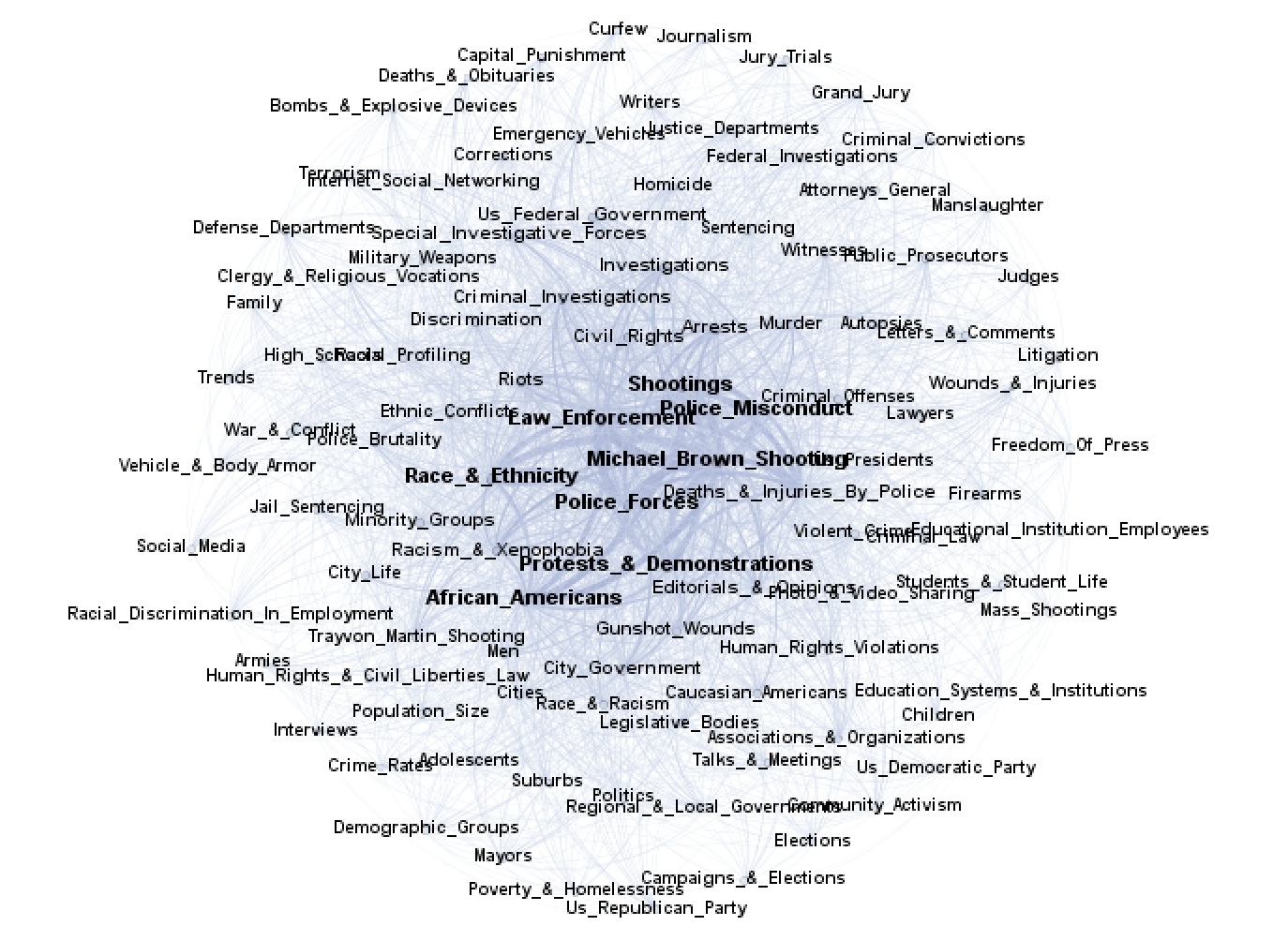} \\
(a) June 2014 & (b) August 2015 \\
\includegraphics[height=6.3cm]{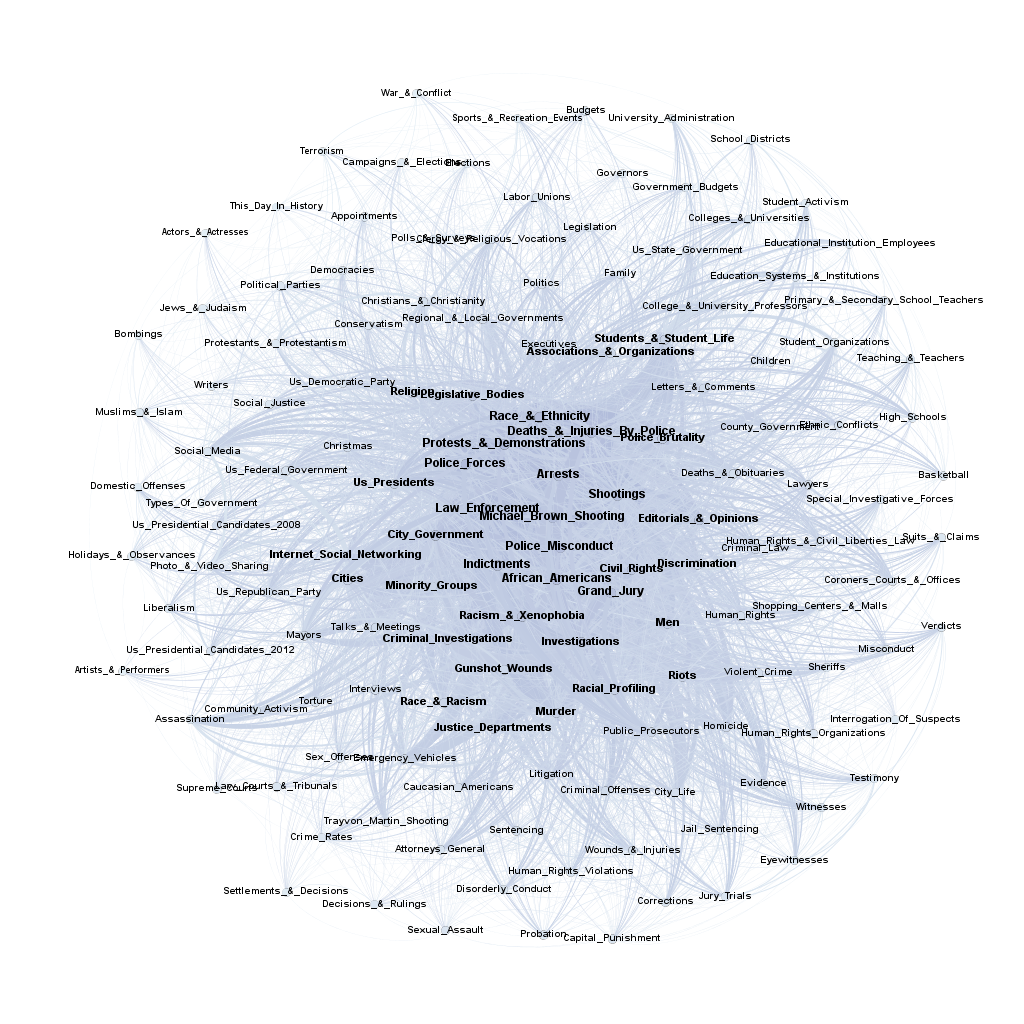} & \includegraphics[height=6.3cm]{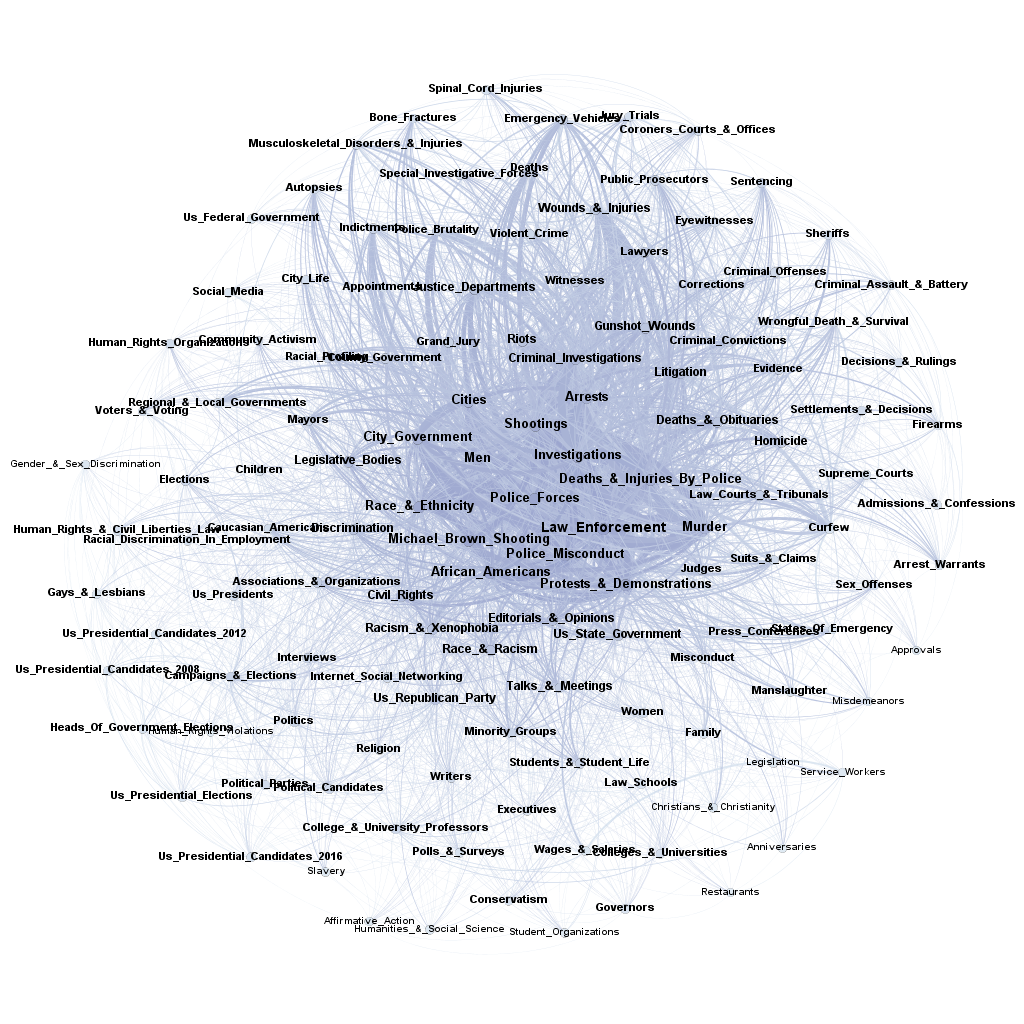} \\
(c) December 2014 & (d) April 2015 \\
\end{tabular}}
\caption{Semantic networks of news articles}
\label{fig:networks}
\end{figure*}


\subsection{Language Dimensions Analysis}

Figure~\ref{fig:ldim} shows statistical analysis of language dimensions in tweets and news articles. As mentioned earlier, analytic thinking shows ``writers' ability to observe and express the underlying issues in a society''. 
Moreover, since news articles are more intended to be formal and logical, compared to the posts on social media, the analytic thinking score of news articles are higher than those of tweets, as expected. A similar observation is found for authenticity. The higher score of language authenticity shows that the writer is expressing personal experience. However, we observe a lower authenticity score on both media and social media which may be the result of expressing second-hand experiences. Overall, the authenticity of news articles are generally 
higher than the authenticity of tweets.

Nevertheless, we found it surprising that the Clout scores of 
news articles and the Clout scores of tweets are similar 
with the average scores of 71.77 and 68.82, respectively.
The results suggest that people are confident when they express opinions regarding the social events, namely BLM movement. 
In addition, reposting news articles on Twitter as an evidence can contribute to similar Clout score to news articles.

Low emotional tone, below 50, reflects anxiety, sadness, and hostility from the writers' perspective. As our analysis shows, the values of emotional tone in the news article are generally lower than the emotional tone of tweets. These results coincide with the results of sentiment analysis where the news articles consisted of mostly negative sentiment.

We also compared the changes in emotional tone with the events
related to the BLM movement. As expected, we found that the emotional tone scores in tweets dropped after significant events took place. The drops in emotional values are shown 
(1) in August 2014 after the incidents of Eric Garner and Michael Brown, (2) in December 2014 after the non-indictments of Darren Wilson and Daniel Pantaleo, and, again, (3) in April 2015 after the incident of Walter Scott and Freddie Gray.
We did not find similar observations for news articles.

\subsection{Semantic Networks Analysis of News Articles}

After extracting the meta-data and the semantic networks from the news articles, we used Gephi \cite{bastian2009gephi} to visualize the networks. We chose four networks to review in this section. Figures~\ref{fig:networks}(a) and~\ref{fig:networks}(b) show semantic networks of news articles taken before and after the important events of BLM movement: deaths of Eric Garner and Michael Brown, respectively.
Before these significant incidents, news articles were focused on local news and everyday events and consisted of the terms such as {\em gunshot wounds}, {\em shootings}, {\em firearms} or {\em murder} related to these events. After the incidents, the semantic networks show more spread of topic focus; however, the terms related to the significant events in the society such as 
the names of the victims, e.g., {\em Michael Brown Shooting},
motives and concerns, e.g., {\em Police Misconduct} and {\em race \& ethnicity}, and activities, e.g., {\em Protests \& Demonstrations} as well as terms such as {\em Ethnic Conflicts} and {\em Human Rights Violation} are added to the discourse.

These change in semantic networks suggest that news articles shifted their discussions towards the incidents in more details 
and provided discussion on the racial biases and events taking place in society. The latter observation derives from both public interviews and articles by journalists regarding covering the incidents, e.g. usage of topics related to {\em African American}, {\em Police Misconduct} and {\em Michael Brown Shootings}.
In fact, the first two terms were not found in the network shown in Figure~\ref{fig:networks} (a) even though the movement was already started since 2013. This shows that the BLM movement received more attention in news articles after significant losses (of two African-American) occurred in the US. We argue that raising awareness regarding these underlying issues need to be taken more seriously by traditional media. 

In addition, we analyzed two additional semantic networks during events in December 2014, Figure~\ref{fig:networks}(c), and April 2015, ~\ref{fig:networks}(d). Our analysis shows similar observation to Figure~\ref{fig:networks}(b). Regarding the topic and discourse, after the non-indictments in December 2014, we found topics such as Indictments and Grand Jury added to the network. After the death of Walter Scott in April 2015 we did not find any significant changes in the topics of the news articles and the discourse. Figures ~\ref{fig:networks}(c) and~\ref{fig:networks}(d) more clearly shows the terms discussed in news during the mentioned events. 

Nevertheless, the semantic networks are shown in Figure~\ref{fig:networks} help to validate our choices of news articles dataset. Moreover, based on the presented indexed terms found in the networks we can see that the query used 
for retrieving the news articles were efficient and helped in extracting the news articles related to the topic of this study.



\section{Conclusion}
Social media and media are powerful sources for studying and finding the impact of events on individuals and societies. Activists groups organize social movements to raise awareness about a social justice issue or to engage the public to change or modify an established legislation. With the emergence of the internet and more importantly social media, activists tried to leverage this medium to expand their target audience and recruit more individuals who share the same interest. Traditional media such as news channels and radio spread the movement from another perspective. While the language on social media is more opinionated, traditional media tends to stay more formal and follow logical conventions when talking about political and social events. 

In this paper, we aimed to analyze the language use on both social media, namely Twitter, and traditional media, news articles, to find similarities and differences in discourse regarding significant social movements, namely Black Lives Matter. We compared emotions, sentiments, language dimensions and individualism vs. pluralism on both mediums. Our results show that individuals on Twitter tend to use more ``we'' and ``our'' when a major incident takes place. This show that people in a society connect and at least virtually take part in the movement. In addition, we found that both emotions and sentiments in language were significantly influenced by the major events in both news and Twitter. Nevertheless, we may need to consider more in-depth linguistic and pragmatic features to study the two mediums. 

Findings of this paper highlights the influence of social movement as a stimuli on the attitude and emotion of people. While previous studies found significant impact of external sources such as information sources on people \cite{rezapour2017classification}, the depth and magnitude of social movements yet need to be studied since they have long term effects on communities and groups. while focusing on this problem in our future work, we will also expand this work to study the usage of individualism vs. pluralism in people's everyday language use to better understand emotion, personality and status. 

\section{Acknowledgment}
I would like to thank Professor Jana Diesner and Professor Karrie Karahalios from University of Illinois at Urbana-Champaign for their helpful insights and direction. I also thank Shubhanshu Mishra for his help with getting the Twitter dataset and Yodsawalai Chodpathumwan for her great assistant with this paper. 

\bibliographystyle{plain}
\bibliography{reference.bib}

\end{document}